\documentclass{article}

\usepackage{microtype}
\usepackage{graphicx}
\usepackage{subfigure}
\usepackage{booktabs} 

\usepackage{hyperref}

\usepackage[accepted]{icml2024}

\usepackage{amsmath}
\usepackage{amssymb}
\usepackage{mathtools}
\usepackage{amsthm}

\usepackage[capitalize,noabbrev]{cleveref}

\theoremstyle{plain}
\newtheorem{theorem}{Theorem}[section]

\theoremstyle{definition}
\newtheorem{definition}[theorem]{Definition}

\theoremstyle{remark}
\newtheorem{remark}[theorem]{Remark}

\usepackage[textsize=tiny]{todonotes}
\usepackage{xspace}
\usepackage{adjustbox, pifont}
\usepackage{multicol, multirow}
\usepackage{dsfont}
\usepackage{enumitem}
\usepackage{tcolorbox}
\tcbuselibrary{skins,breakable}
\tcbset{enhanced jigsaw}

\makeatletter
\newcommand{\thickhline}{%
    \noalign {\ifnum 0=`}\fi \hrule height 1.3pt
    \futurelet \reserved@a \@xhline
}

\renewcommand{\epsilon}{\varepsilon}

\newenvironment{tbox}{\begin{tcolorbox}[
		enlarge top by=5pt,
		enlarge bottom by=5pt,
		 breakable,
		 boxsep=0pt,
                  left=4pt,
                  right=4pt,
                  top=10pt,
                  arc=0pt,
                  boxrule=1pt,toprule=1pt,
                  colback=white
                  ]
	}
{\end{tcolorbox}}

\DeclareMathOperator*{\argmin}{argmin}

\newcommand{\cpd}{\ensuremath{\textnormal{\textsc{Rio-CPD}}}\xspace}

\newcommand{\spdn}{\ensuremath{\mathcal{S}_+^n}\xspace}

\icmltitlerunning{\textbf{RIO-CPD}: Correlation-aware Online Change Point Detection}

\begin{document}

\twocolumn[
\icmltitle{\textbf{RIO-CPD}: A Riemannian Geometric Method for \\Correlation-aware Online Change Point Detection}



\begin{icmlauthorlist}
\icmlauthor{Chengyuan Deng}{yyy}
\icmlauthor{Zhengzhang Chen}{comp}
\icmlauthor{Xujiang Zhao}{comp}
\icmlauthor{Haoyu Wang}{comp}\\
\icmlauthor{Junxiang Wang}{comp}
\icmlauthor{Haifeng Chen}{comp}
\icmlauthor{Jie Gao}{yyy}
\end{icmlauthorlist}

\icmlaffiliation{yyy}{Rutgers University}

\icmlaffiliation{comp}{NEC Laboratories America}
\icmlcorrespondingauthor{Zhengzhang Chen}{zchen@nec-labs.com}

\icmlkeywords{Change Point Detection, Riemannian Metrics, Online Learning, CUSUM, Correlation-awareness}

\vskip 0.3in
]



\printAffiliationsAndNotice{Work done during an internship at NEC Laboratories America.} 

\begin{abstract}
Change point detection aims to identify abrupt shifts occurring at multiple points within a data sequence. This task becomes particularly challenging in the online setting, where different types of changes can occur, including shifts in both the marginal and joint distributions of the data. In this paper, we address these challenges by tracking the Riemannian geometry of correlation matrices, allowing Riemannian metrics to compute the geodesic distance as an accurate measure of correlation dynamics. 
We introduce \textsc{Rio-CPD}, a non-parametric, correlation-aware online change point detection framework that integrates the Riemannian geometry of the manifold of symmetric positive definite matrices with the cumulative sum (CUSUM) statistic for detecting change points. \textsc{Rio-CPD} employs a novel CUSUM design by computing the geodesic distance between current observations and the Fr\'echet mean of prior observations. With appropriate choices of Riemannian metrics, \textsc{Rio-CPD} offers a simple yet effective and computationally efficient algorithm. Experimental results on both synthetic and real-world datasets demonstrate that \textsc{Rio-CPD} outperforms existing methods on detection accuracy, average detection delay and efficiency.

\end{abstract}

\section{Introduction}
\label{sec:intro}


The task of change point detection (CPD)~\cite{page1954continuous} seeks to identify abrupt distributional changes in the temporal evolution of a system through noisy observations. Localizing these changes helps isolate and interpret different time series patterns and has crucial implications for system safety and reliability. CPD has applications across domains such as climatology\cite{reeves2007review, gallagher2013changepoint}, finance~\cite{pepelyshev2015real, lavielle2007adaptive}, and healthcare~\cite{yang2006adaptive}. CPD can be conducted in an offline or online manner, with the latter being more challenging due to its real-time nature, requiring minimal delay in detecting changes. We focus on unsupervised \emph{online} CPD in a discrete setting, where multivariate time series are streamed, multiple change points may occur, and no labels are revealed throughout the process.


Despite significant progress in online CPD techniques in recent years~\cite{keriven2020newma, alanqary2021change, caldarelli2022adaptive, li2024automatic, wu2023score}, efficiently capturing changes in various patterns, such as those in the marginal distribution (\textit{e.g.}, independent magnitude) and joint distribution (\textit{e.g.}, correlations between covariates), remains challenging. Correlation-aware CPD methods~\cite{lavielle1999detection, barnett2016change, cabrieto2018testing, cabrieto2017detecting, zhang2020correlation, yu2023dynamic} have gained attention due to their practical impact in fields such as behavioral science~\cite{mauss2005tie} and root cause analysis in AIOps~\cite{wang2023incremental, wang2023interdependent, chen2024trigger}. However, these methods were not originally designed for online use. The reason comes in two-fold: first, extracting correlations for CPD in multivariate time series often involves complex techniques like Graph Neural Networks or probabilistic graphical models, which are inherently challenging to adapt to an online setting. Second, these methods typically require extended processing time, making them inefficient for real-time applications. As a result, despite the pressing need, there has been limited work specifically tailored for correlation-aware online CPD.

\begin{figure*}[h]
    \centering
    \includegraphics[width=\textwidth]{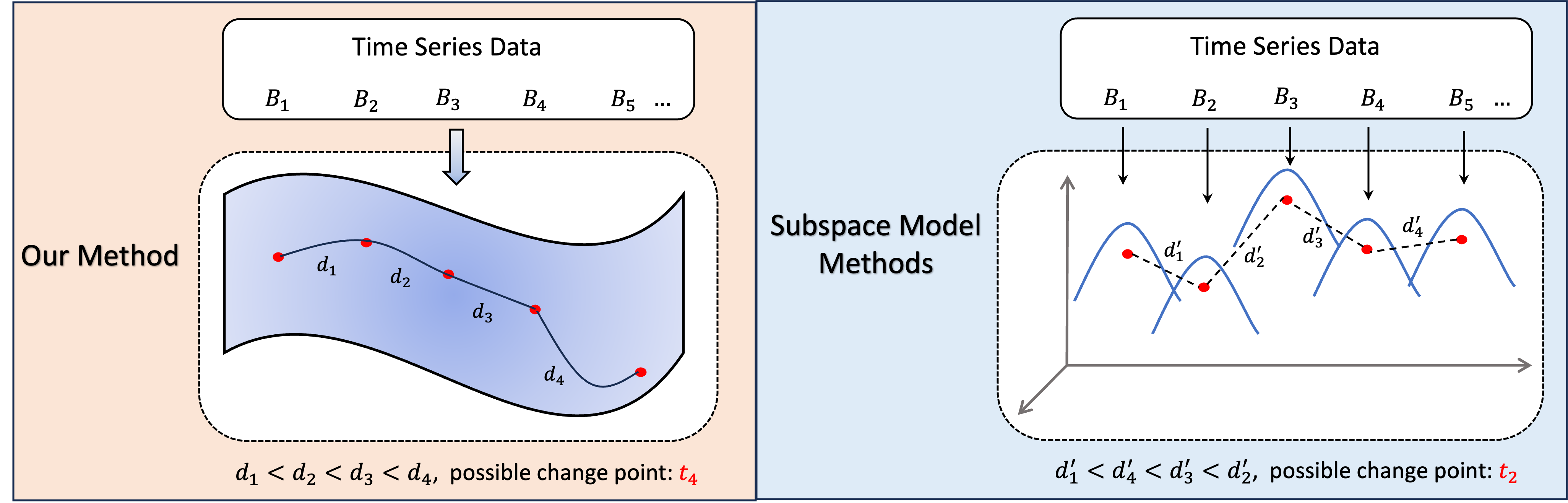}
    \vspace{-5mm}
    \caption{Comparison between \cpd and subspace model methods. $B_i$ represents the $i$-th batch of time series data, and $d_i$ denotes the distances between neighboring batches within the subspace.}
    \label{fig:teaser}
\end{figure*}



On the other hand, a significant portion of existing work on online CPD falls under the category of \emph{Subspace Models}. These approaches estimate a subspace for each batch of time series data and then use a metric to compute the distance between consecutive batches, as illustrated in \Cref{fig:teaser}. A change point is detected if there is an abrupt change in this distance. Examples of subspace estimation techniques include singular spectrum analysis~\cite{alanqary2021change}, state space models~\cite{fuh2020asymptotically}, dimension reduction~\cite{jiao2018subspace}, and refined metric design~\cite{costa2006single, keriven2020newma, dubey2020frechet}. Although subspace models are relatively easy to adapt to online settings, they can suffer from accuracy issues due to distortions in subspace approximation and efficiency challenges from high computational costs.

In this study, we extend the subspace model for correlation-aware online CPD, addressing the challenges discussed above. We introduce a non-parametric framework called \cpd, a \underline{Ri}emannian geometry-based method for correlation-aware \underline{O}nline \underline{C}hange \underline{P}oint \underline{D}etection.




At a schematic level, \cpd is inspired by the natural observation that Pearson correlation matrices are symmetric positive semi-definite, hence can be characterized by Riemannian geometry and all fall into one subspace of Riemannian manifold. The distance between the correlations of two batches of time series data corresponds to the geodesic between two points on the Riemannian manifold, which can be computed using a specific Riemannian metric. These distances are then used to construct the cumulative sum (CUSUM) statistic, which performs sequential hypothesis testing to determine whether a time step is a change point. The \cpd framework is flexible with respect to the choice of Riemannian metrics and CUSUM statistic construction, although there may be trade-offs. We employ the Log-Euclidean and Log-Cholesky metrics, which avoid Riemannian optimization in CUSUM statistic computation, thereby ensuring efficiency. Details of our design are provided in \Cref{sec:method}.

We provide a high-level comparison of \cpd with the subspace model in \Cref{fig:teaser}. \cpd addresses the limitations of subspace models by (1) directly operating on the Riemannian manifold of correlation matrices rather than an estimated subspace, which may introduce distortion, and (2) bypassing the time-consuming subspace learning process. Consequently, the \cpd framework enhances the subspace model with improved accuracy and efficiency.

Our contributions are summarized as follows:
\begin{itemize}
    \item \textit{\textbf{Problem}}: We tackle the problem of correlation-aware online change point detection, aiming to accurately and efficiently identify changes in both individual variables and correlations within multivariate time series data. Existing methods either lack the ability to handle correlations or are unsuitable for real-time detection, motivating the need for a new approach.
    \item \textit{\textbf{Framework}}: We propose \cpd, a non-parametric framework that leverages the Riemannian geometry of correlation matrices and the CUSUM procedure to detect change points. \cpd is capable of detecting changes in both independent and correlated patterns while being highly efficient and flexible in choosing Riemannian metrics to balance performance trade-offs.
   \item \textit{\textbf{Evaluations}}: We conduct extensive experiments on both synthetic and real-world datasets to validate the effectiveness of our approach. The results show that \cpd significantly outperforms state-of-the-art methods in terms of detection accuracy and efficiency.

\end{itemize}

\section{Preliminary}
\subsection{Problem Statement}
Let $X(t) = [X_1(t), \dots, X_m(t)] \in \mathbb{R}^m$ be an observation of a discrete multivariate time series at time index $t$, where $t \in [T]$, and denote $f_i:\mathbb{Z}^+ \rightarrow \mathbb{R}$ as the latent distribution of the $i$-th time series, such that the observation at time $t$ take the form of $X(t) = f(t) + e(t)$, where $f(t) = [f_1(t), \dots ,f_m(t)]$ and $e(t)$ is a zero-mean i.i.d. random variable representing the noise. In particular, the change point detection task aims to find all $\tau \in [T]$ such that:

\vspace{-2mm}
\begin{align*}
    X(t) =
    \begin{cases}
        f(t) + e(t), &t<\tau \\
        f'(t) + e(t), &t \geq \tau
    \end{cases}
\end{align*}
for some functions $f \neq f'$. Another perspective of the CPD problem is to perform sequential hypothesis testing at each time step $t$ such that one of the following is accepted:
\begin{align*}
    &H_0: \mathbb{E}[X(t)] = f(t) \\
    &H_1: \mathbb{E}[X(t)] = f'(t)
\end{align*}

Let $\hat{\tau} = \inf \{ t |  H_1 \text{ is accepted at }t\}$, for any CPD algorithm, an essential property is to identify the change point promptly, i.e. $\hat{\tau} > \tau$ and $\hat{\tau} - \tau$ is small.


\subsection{Standard Notions in Riemannian Geometry}

A manifold $\mathcal{M}$ is a topological space that is locally diffeomorphic to Euclidean space. To measure geometric properties on such a space, a Riemannian manifold is defined as a manifold equipped with a Riemannian metric, formally,

\begin{definition}[Riemannian Manifold]
    \label{def:rie-manifold}
    A Riemannian manifold $(\mathcal{M},g)$ is a smoothed manifold $\mathcal{M}$ endowed with a smoothly varying family of inner products $g_x:T_x\mathcal{M} \times T_x\mathcal{M} \rightarrow \mathbb{R}$, where $T_x\mathcal{M}$ is the tangent space of $\mathcal{M}$ at $x \in \mathcal{M}$.
\end{definition}


The geodesic of a Riemannian manifold can be defined using the inner product. Specifically, for two points \(p, q \in \mathcal{M}\), the geodesic distance between them follows:
\begin{align}
d_{\mathcal{M}}(p, q) = \int_0^1 \sqrt{g(\gamma'(t), \gamma'(t))} \, dt,
\end{align}
where \(\gamma: [0, 1] \rightarrow \mathcal{M}\) is a smooth curve such that \(\gamma(0) = p\) and \(\gamma(1) = q\).


To facilitate computations on such manifolds, the exponential and logarithm maps play a crucial role by linking the geometry of a Lie group with its Lie algebra. The Lie algebra can be seen as the tangent space of the Lie group at the identity element, providing a local linear approximation of the manifold. In our context, these maps establish a one-to-one correspondence between symmetric positive definite (SPD) matrices and a vector space structure, ensuring that geodesics are well-defined and can be computed effectively.


\begin{definition}[Exponential and Logarithm Map of Matrices]
    \label{def:exp-log-map}
    Given a matrix $A$, the exponential map is defined by $\exp(A) := \sum_{k=0}^\infty A^k/k!$,  while the logarithm map, denoted as $\log(A)$, serves as its inverse.
\end{definition}

Finally, we introduce the Fr\'echet mean of a set of matrices within a given metric space.
\begin{definition}[Fr\'echet mean]
    \label{def:frechet-mean}
    Let $(\mathcal{M},d)$ be a complete metric space, and let $P_1, \dots P_n$ be points in $\mathcal{M}$. The Fr\'echet mean of $P_1, \dots P_n$ is defined as:
    \begin{align}
        \sigma_\mathcal{M} = \argmin_{x \in \mathcal{M}}\sum_{i=1}^n d^2(x,P_i)
    \end{align}
\end{definition}

\section{Methodology}
\label{sec:method}

Our approach is motivated by the observation that, in practice\footnote{Correlation matrices are positive semi-definite, but under reasonable assumptions, such as linear independence of data columns, they are positive definite.}, correlation matrices lie in $\spdn$, which exhibits Riemannian structures~\cite{arsigny2007geometric}. This allows us to apply a Riemannian metric to correlation matrices with minimal distortion. Based on this metric, the \cpd framework proceeds in three stages.

First, the Riemannian metric is used to track the distances between consecutive correlation matrices until a change point is identified. Next, we construct the CUSUM statistics by measuring the distance between the current correlation matrix and the ``centroid'' of the sub-manifold representing the collection of past correlation matrices. If a change point is present, this distance will be significantly large, indicating that the current matrix should belong to the cluster representing the post-change distribution, and is an ``outlier'' to the current one. Finally, the third stage involves applying the CUSUM sequential hypothesis test to detect and report a change point as soon as it occurs.


It is important to note that constructing the CUSUM statistics is non-trivial, as it is closely related to the choice of Riemannian metrics. In general, the centroid of a Riemannian manifold, represented by the Fr\'echet mean, may not have a closed-form solution and often requires approximation through optimization. However, with the use of two specific Riemannian metrics, known as the Log-Euclidean and Log-Cholesky metrics, the Fr\'echet mean has a closed-form solution. This property allows \cpd to bypass the complexities of Riemannian optimization.

\begin{figure*}[h!]
    \centering
    \includegraphics[width=\textwidth]{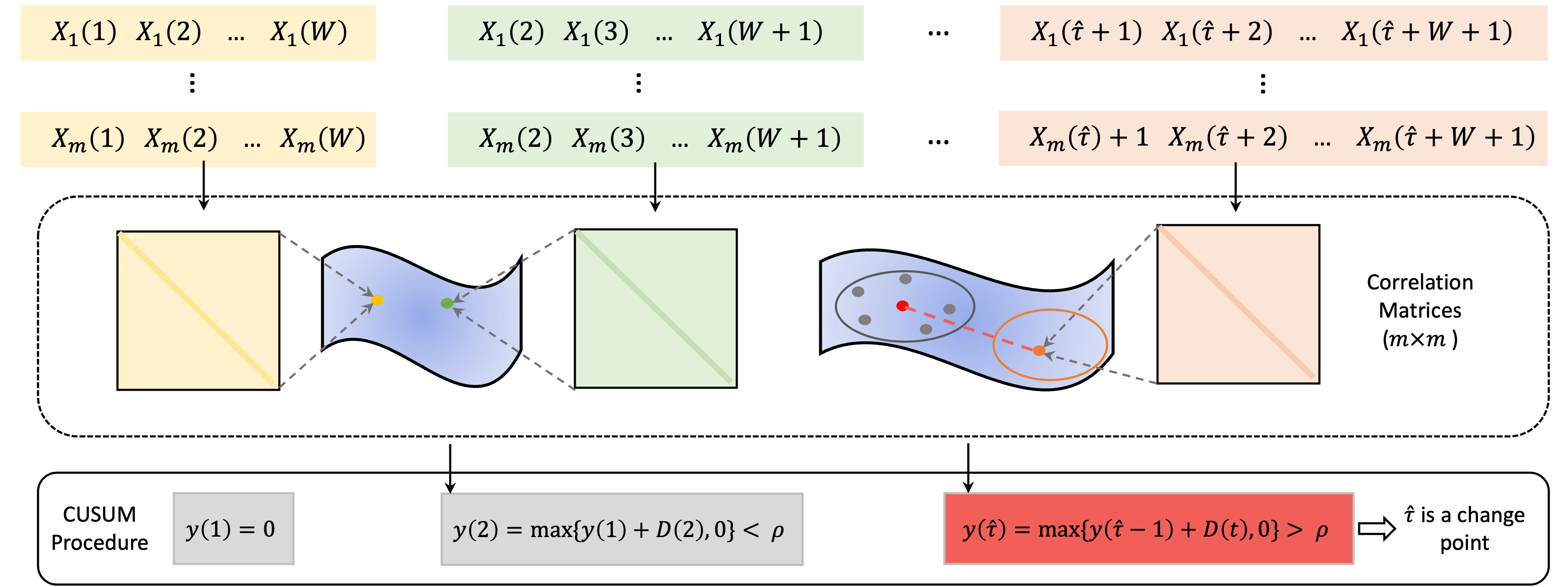}
    \vspace{-5mm}
    \caption{The overview of the proposed \cpd framework. First, a correlation matrix is constructed for each batch of data within the sliding window. Next, the distance between the current observation and the centroid of previous ones is calculated. A change point is likely to belong to a different cluster, resulting in a larger distance from the centroid. Finally, we compute the CUSUM statistics using this distance and perform a hypothesis test to detect the change point.}
    \label{fig:alg}
\end{figure*}

In the following sections, we introduce the CUSUM procedure and the Riemannian metrics, and conclude by presenting the entire algorithm in \Cref{subsec:alg}. An illustration of the \cpd framework is provided in \Cref{fig:alg}.

\subsection{CUSUM Procedure}
\label{subsec:CUSUM}

The CUSUM statistic is a measure of the likelihood that a given timestamp represents a change point, and it has been widely used in hypothesis testing. To construct the CUSUM statistic at timestamp $t$, a detection score is required, which typically evaluates the ``distance'' between the observation at timestamp $t$ and the distribution of the time series observed so far. Formally, 

\begin{definition}[CUSUM Statistic]
    \label{def:cusum}
    Given an observation $X(t)$ at time step $t$ and a detection score $D(t)$, the CUSUM statistic of time series $\{X(1), \dots, X(t)\}$ is defined as $y(t) = \max_{1\leq i \leq t}\sum_{j=i}^t D(j)$.
\end{definition}


It is well-known from the literature~\cite{page1954continuous} that \Cref{def:cusum} can alternatively be expressed as $y(t) = \max\{y(t-1)+D(t), 0\}$, where $y(0) = 0$. This recursive formulation improves the efficiency of testing by utilizing only the previous step's history. The CUSUM procedure proceeds as follows: At each timestamp $t$, we perform the hypothesis test:

\vspace{-7mm}
\begin{align*}
    H(t) = H_{\mathds{1}_{y(t) \geq \rho}}, &\text{ where } \\
& H_0: t \text{ is not a change point,} \\
& H_1: t \text{ is a change point.}
\end{align*}
Note that $\rho > 0$ is a threshold. Therefore, the set of change points is given by ${t \mid y(t) \geq \rho}$.


Although the CUSUM procedure may seem straightforward for change point detection, designing the detection score $D(t)$ is often challenging and plays a critical role in the overall performance of the method. In parametric settings, $D(t)$ can be defined as the log-likelihood ratio between two distributions. However, this approach requires knowledge of the post-change distribution, which is often unavailable in practice and limits its applicability to general scenarios.

In the following sections, we explain how \cpd constructs $D(t)$. A key requirement for $D(t)$ is its ability to effectively differentiate between change points and regular timestamps, which necessitates that $\mathbb{E}[D(t)|H_0] < 0$ and $\mathbb{E}[D(t)|H_1] > 0$.


\subsection{Log-Euclidean Metric}
We first introduce the Lie group structure of \spdn, which enables a bi-invariant Riemannian metric derived from any inner product between \spdn and its tangent space. Given two matrices $P_1, P_2 \in \spdn$, we define the logarithm addition and multiplication, with the exponential and logarithm map (see \Cref{def:exp-log-map}).
\begin{align*}
    P_1 \oplus P_2 &= \exp(\log P_1 + \log P_2) \\
    \lambda \odot P_1 &= \exp(\lambda \cdot P_1)
\end{align*}
where $\lambda$ is a real number. 

The matrix logarithm for Log-Euclidean is defined as:
\begin{align*}
    \phi_{LE}(P_1) = U\ln(\Sigma)U^T
\end{align*}
where $P_1 = U\Sigma U^T$ is the eigenvalue decomposition. 

The geodesic distance under the Log-Euclidean metric is given by:
\begin{align}
    d_{LE}(P_1, P_2) = \|\phi_{LE}(P_1) - \phi_{LE}(P_2) \|_F
\end{align}

While the computation of the Log-Euclidean metric can be theoretically demanding due to the logarithm map, there are techniques available to reduce computation time. One notable advantage of the Log-Euclidean metric is that it provides a closed-form solution for the Fr\'echet mean of SPD matrices, which can be regarded as the ``centroid'' of the $\spdn$ geometry. Similar to the centroid in Euclidean space (\textit{e.g.}, the $k$-means objective), the Fr\'echet mean minimizes the squared error of the geodesic distances to all SPD matrices (\Cref{def:frechet-mean}).


The Fr\'echet mean of Log-Euclidean geometry (also known as Log-Euclidean mean) is given by~\cite{arsigny2007geometric}:

\vspace{-5mm}
\begin{align}
    \label{eq:log-euc-frechet}
    \sigma_{LE}(P_1, \dots, P_n) = \exp\big( \frac{1}{n} \sum_{i=1}^n\phi_{LE}(P_i)\big)
\end{align}


It is worth noting that the Log-Euclidean mean is a natural generalization of the geometric mean. If $P_i$ are positive real numbers, their geometric mean follows the same formula. The Log-Euclidean and Log-Cholesky metrics are the only Riemannian metrics that admit a closed-form solution for the Fr\'echet mean. For other metrics, the Fr\'echet mean may not even be unique.

\subsection{Log-Cholesky Metric}
The Log-Euclidean and Log-Cholesky metrics~\cite{lin2019riemannian} share several similarities and, at a high level, are topologically equivalent because $\mathcal{L}^n \simeq \mathcal{S}^n \simeq \mathbb{R}^{n(n+1)/2}$ \footnote{Notation: $\mathcal{L}^n$ denotes the Cholesky space of dimension $n$, and $\simeq$ represents homotopic equivalence. These notations are used only once here.}. The Log-Euclidean metric is obtained through the matrix logarithm: $\spdn \rightarrow \mathcal{S}^n$, while the Log-Cholesky metric is based on $\spdn \rightarrow \mathcal{L}^n$. The definition of the Log-Cholesky metric begins with the Cholesky decomposition, which provides a unique representation of any Hermitian and positive definite matrix $A$ in the form $LL^T$, where $L$ is a lower triangular matrix and $L^T$ is its upper triangular counterpart. The matrix logarithm for the Log-Cholesky geometry is defined as:
\vspace{-1mm}
\begin{align*}
    \phi_{LC}(P_1) = \lfloor L \rfloor+ \ln(\mathbb{D}(L))
\end{align*}
where $\lfloor L \rfloor$ is the strictly lower triangular part of $L$, and $\mathbb{D}(L)$ denotes the diagonal matrix extracted from $L$, with all other entries set to zero.

The geodesic distance under Log-Cholesky metric is defined as:
\begin{align}
\begin{split}
    d_{LC}(P_1, P_2) & = \|\phi_{LC}(P_1) - \phi_{LC}(P_2) \|_F \\
    & = \Big(\|\lfloor L_1 \rfloor -  \lfloor L_2 \rfloor \|_F^2 \\
    &  \quad  + \|\ln\big(\mathbb{D}(L_1)\big)-\ln \big(\mathbb{D}(L_2)\big) \|_F^2\Big)^{1/2}
\end{split}
\end{align}
where $P_1 = L_1 L_1^T$ and $P_2 = L_2 L_2^T$.

The Log-Cholesky metric also has a convenient closed-form expression for its Fr\'echet mean over $n$ matrices $P_1, \dots, P_n \in \spdn$:
\vspace{-1mm}
\begin{align*}
    \sigma_{LC}(P_1, \dots, P_n) = \sigma_{L}(L_1, \dots, L_n)\cdot \sigma_{L}(L_1, \dots, L_n)^T
\end{align*}
where $P_i$ has Cholesky decomposition $L_i L_i^T$ and $\sigma_L$ is the Fr\'echet mean of Cholesky space $\mathcal{L}^n_+$:

\vspace{-3mm}
\begin{align}
\label{eq:log-cho-frechet}
    \sigma_L(P_1, \dots, P_n) = \frac{1}{n}\sum_{i=1}^n \lfloor L_i \rfloor + \exp\big( \frac{1}{n} \sum_{i=1}^n \log \mathbb{D}(L_i)\big)
\end{align}

\subsection{RIO-CPD Algorithm}
\label{subsec:alg}
\cpd requires one parameter to be specified in advance: the size of the sliding window $W$. The algorithm proceeds in three steps at each time step $t$: (1) Using the time series data from $t$ to $t+W-1$, construct the correlation matrix $\mathbf{B}_t$. (2) Monitor the geodesic distance between $\mathbf{B}_t$ and $\sigma_{t-1}$, the Fr\'echet mean of the collection of previous correlation matrices. (3) Compute the CUSUM statistic based on this distance and perform hypothesis testing. If the test does not detect a change point, proceed to $t+1$; otherwise, report a change point at $t$ and repeat the process for the next batch of streaming data.

\textbf{Step 1.} Transform observations in the sliding window into correlation matrices. Suppose at time $t$ we have $\mathbf{X}_t = [X(t), \dots, X(t + W -1)]$, then construct

 \vspace{-2mm}
\begin{align*}
    \mathbf{B}_t = \tilde{\mathbf{X}}_t \cdot \tilde{\mathbf{X}}^T_t, \text{where } \tilde{\mathbf{X}}_t = \frac{\textbf{X}_t - \mathbb{E}[\textbf{X}_t]}{Var(\textbf{X}_t)} \text{ is normalized.}
\end{align*}

\textbf{Step 2.} Let $\mathfrak{g}$ be either the Log-Euclidean or Log-Cholesky metric. We compute $\sigma_{t-1}^\mathfrak{g}$, the Fr\'echet mean of $\mathbf{B}_1, \dots, \mathbf{B}_{t-1}$ using \Cref{eq:log-euc-frechet} or \Cref{eq:log-cho-frechet}. Next, we calculate the distance from $\mathbf{B}_t$ to $\sigma_{t-1}^\mathfrak{g}$, \textit{i.e.}, $d_t = \mathfrak{g}(\mathbf{B}_t, \sigma_{t-1}^\mathfrak{g})$. \footnote{Here we consider a 1-lag. If the observations are sampled at high frequency, the algorithm can use $\mathbf{B}_{1}, \mathbf{B}_{1+L}, \mathbf{B}_{1+2L}, \dots$, where $L \geq 1$ becomes a parameter.}

\textbf{Step 3.} Construct the detection score $D(t)$ required for CUSUM. We define $D(t)$ as $d_t$ minus $r_{t-1}$, the radius of the subspace formed by $\mathbf{B}_1, \dots, \mathbf{B}_{t-1}$, i.e.


 \vspace{-2mm}
\begin{align*}
    D(t) = d_t - r_{t-1} =  \mathfrak{g}(\mathbf{B}_t, \sigma_{t-1}^\mathfrak{g}) - \max_{i\in[t-1]}\mathfrak{g}(\mathbf{B}_i, \sigma_{t-1}^\mathfrak{g})
\end{align*}



With the detection score $D(t)$, the CUSUM test iteratively computes $y(t)$ for $t > W$, starting with $y(W) = 0$, until a change point is detected, \textit{i.e.}, $y(t) > \rho$ for a given threshold $\rho$. Specifically,
\begin{align*}
    \hat{\tau} = \inf_{t>W}\{t \mid y(t) > \rho \}.
\end{align*}

The procedure can be extended to handle multiple change points by restarting the process at $\hat{\tau}+1$ after detecting a change point $\hat{\tau}$, with a new base correlation matrix $\mathbf{B}_{\hat{\tau} +1}$. We provide further insights into our algorithm and discuss its computational complexity in the remarks below.



\begin{remark}
    The design of the detection score $D(t)$ is inspired by clustering and incorporates strong geometric insights. Essentially, \cpd first projects the correlation dynamics into the Riemannian manifold, if the input data follows a stable distribution, the projection would become a cluster of points in the Riemannian manifold. Therefore, a change in distribution would lead to a shift of the cluster. \cpd identifies $t$ as a change point if $\mathbf{B}_t$ is an outlier from the cluster formed by $\mathbf{B}_1, \dots, \mathbf{B}_{t-1}$. To detect an outlier, we calculate the difference between the current distance and the ``radius'' of the cluster. If this difference exceeds a certain threshold, it is likely that $\mathbf{B}_t$ is an outlier, indicating a change point.
\end{remark}

\begin{remark}
    The primary computational bottleneck for our algorithm arises from eigen decomposition (Log-Euclidean) and Cholesky decomposition (Log-Cholesky), both of which are performed with $O(m^3)$ time complexity for $m$-dimensional symmetric matrices. In theory, the exponent can be reduced to a constant close to $\omega$ (the fast matrix multiplication constant). Note that $m$ is the number of time series in our setting and is typically a constant, making the algorithm highly efficient in practice.
\end{remark}

\section{Experiments}

In this section, we evaluate the performance of \cpd in terms of accuracy and efficiency. 
We begin by evaluating general scenarios in which the multivariate time series are derived from a dynamic system, with various types of change points. We then consider the task of human action recognition as a concrete application of change point detection and present the corresponding results. 
\label{sec:exp}
\subsection{Setup}

\paragraph{Datasets.} 

We evaluate our model \cpd using both synthetic and real-world datasets. For synthetic data, we consider three datasets derived from a particle-spring system~\cite{zhang2020correlation}, which features five particles moving within a rectangular space. The particles are randomly connected by invisible springs, and their movement follows physical laws such as Newton's and Hooke's laws. The \emph{Connection} dataset exhibits correlation changes, with change points occurring when the spring connections alter. In contrast, the \emph{Speed} and \emph{Location} datasets show change points related to the particles' speed and location, respectively. For \emph{real-world data}, we use several benchmark datasets commonly compared in prior works, including \emph{Beedance}\footnote{\href{https://sites.cc.gatech.edu/~borg/ijcv_psslds/}{https://sites.cc.gatech.edu/~borg/ijcv\_psslds/}} and \emph{HASC}\footnote{\href{http://hasc.jp/hc2011/}{http://hasc.jp/hc2011/}}, which do not necessarily exhibit correlation-based change points. Additionally, we include the \emph{Product Review Microservice} dataset~\cite{zheng2024lemma}, which contains change points due to system irregularities. In this dataset, the system performance metrics are correlated because of the underlying system architecture. The basic statistics for all datasets are presented in~\Cref{tab:dataset}, and further details are available in \Cref{apdx:data}.

\vspace{-2.5mm}
\begin{table}[h!]
\centering
\renewcommand{\arraystretch}{1.3}
\caption{\small Dataset statistics. \#CP represents the number of change points, and \#TS denotes the number of time series. A checkmark in the Correlation column indicates that the change point is due to correlation changes.}
\vspace{0.2em}
\label{tab:dataset}
\begin{adjustbox}{width=0.5\textwidth}
\begin{tabular}{c|ccccc} 
\thickhline
   Dataset &  Length & \#CP & \#TS & Dimension & Correlation\\ 
\hline
  Connection & 100 & 1 & 50 & $\mathbb{R}^5$ & \ding{51} \\
  Speed & 100 & 1 &  50 & $\mathbb{R}^5$ & \ding{55} \\
  Location & 100 & 1 & 50 & $\mathbb{R}^5$ & \ding{55} \\
 \hline 
  Microservice & 1548-1767 & 8 & 4 & $\mathbb{R}^6$ & \ding{51}\\
  Beedance & 608-1124 & 117 & 6 & $\mathbb{R}^3$ & \ding{55}\\
  HASC & 11738-12000 & 196 & 18 & $\mathbb{R}^3$ & \ding{55}\\
 
\thickhline
\end{tabular}
\end{adjustbox}
\end{table}



\paragraph{Baselines.} We compare \cpd with four baselines representing different families of online change point detection (CPD) methods: \textbf{KL-CPD}~\cite{chang2019kernel}, which extends the kernel two-sample test and optimizes a lower bound of the test power via an auxiliary generative model; \textbf{BOCPDMS}, the Spatial-temporal Bayesian Online CPD~\cite{knoblauch2018spatio}, which augments the vanilla Bayesian CPD with Bayesian vector autoregressions; \textbf{MSSA-CPD}~\cite{alanqary2021change}, which employs CUSUM statistics based on subspace estimation via multivariate Singular Spectrum Analysis; and \textbf{Contra-CPD}~\cite{puchkin2023contrastive}, which designs a test statistic that extends the concept of maximizing the discrepancy between pre- and post-change distributions. For Contra-CPD, we select the polynomial function family among its variants.

\paragraph{Evaluation metrics.} We evaluate detection performance primarily using the F1 score, which is crucial for assessing accuracy, as detection can be treated as a binary classification problem. The second metric is detection delay, which is particularly important in practice when delays can have significant consequences. The delay is reported by counting the number of time steps. We also report running time to assess efficiency.

\subsection{Detection Performance}

We first evaluate the accuracy of \cpd on both synthetic and real-world datasets and report the F1 score under two settings: \emph{Default}, which uses the vanilla parameter initialization, and \emph{Best}, which involves a fine-tuning process. A change point is considered successfully detected if it falls within the sliding window reported by the algorithm. For baseline methods without a sliding window, we omit the error caused by the value of $W$ to ensure a fair comparison. 

\vspace{-2mm}
\begin{table}[h!]
\centering
\small
\renewcommand{\arraystretch}{1.2}
\caption{F1 score on real and synthetic datasets. A higher score indicates better performance.}
\label{tab:res-syn}
\begin{adjustbox}{width=0.5\textwidth}
\begin{tabular}{c|cccccc} 
\thickhline
  \multirow{2}{*}{Algorithm} & \multicolumn{2}{c}{Microservice} & \multicolumn{2}{c}{Beedance} &  \multicolumn{2}{c}{HASC} \\
   & Default & Best & Default & Best & Default & Best \\
\hline
  KL-CPD & 0 & 0 & 0.092 & 0.167 & 0.078 & 0.204 \\
  BOCPDMS & 0.061 & 0.109 & 0.092 & 0.167 & 0.078 & 0.204 \\
  MSSA-CPD & 0.154 & 0.308 & 0.500 & \textbf{0.659} & 0.177 & 0.327\\
  Contra-CPD & 0.122 & 0.258 & 0.167 & 0.293 & 0.151 & 0.239\\
  $\mathbf{\cpd}$ (\textbf{LE}) & 0.778 & 0.875 & 0.518 & 0.625 & 0.320 & 0.345\\
  $\mathbf{\cpd}$ (\textbf{LC}) & \textbf{0.933} & \textbf{0.933}  & \textbf{0.535} & 0.643 & \textbf{0.360} & \textbf{0.463}\\
 
\thickhline
  & \multicolumn{2}{c}{Connection} & \multicolumn{2}{c}{Speed} &  \multicolumn{2}{c}{Location} \\
   & Default & Best & Default & Best & Default & Best  \\
\hline
  KL-CPD & 0.087 & 0.120 & 0.155 & 0.263 & 0.052 & 0.199 \\
  BOCPDMS & 0.031 & 0.096 & 0 & 0.114 & 0 & 0.043 \\
  MSSA-CPD & 0.179 & 0.330 & 0.292 & 0.485 & 0.153 & 0.308\\
  Contra-CPD  & 0.268 & 0.304 & 0.315 & 0.396 & 0.283 & 0.342\\
  $\mathbf{\cpd}$ (\textbf{LE})& 0.446 & 0.496 & 0.412 & \textbf{0.510} & 0.378 & \textbf{0.493}\\
  $\mathbf{\cpd}$ (\textbf{LC})& \textbf{0.494} & \textbf{0.511} & \textbf{0.473} & 0.500 & \textbf{0.459} & 0.482\\
  \thickhline
\end{tabular}
\end{adjustbox}
\end{table}

\begin{figure}[h]
    \centering
    \includegraphics[width=\linewidth]{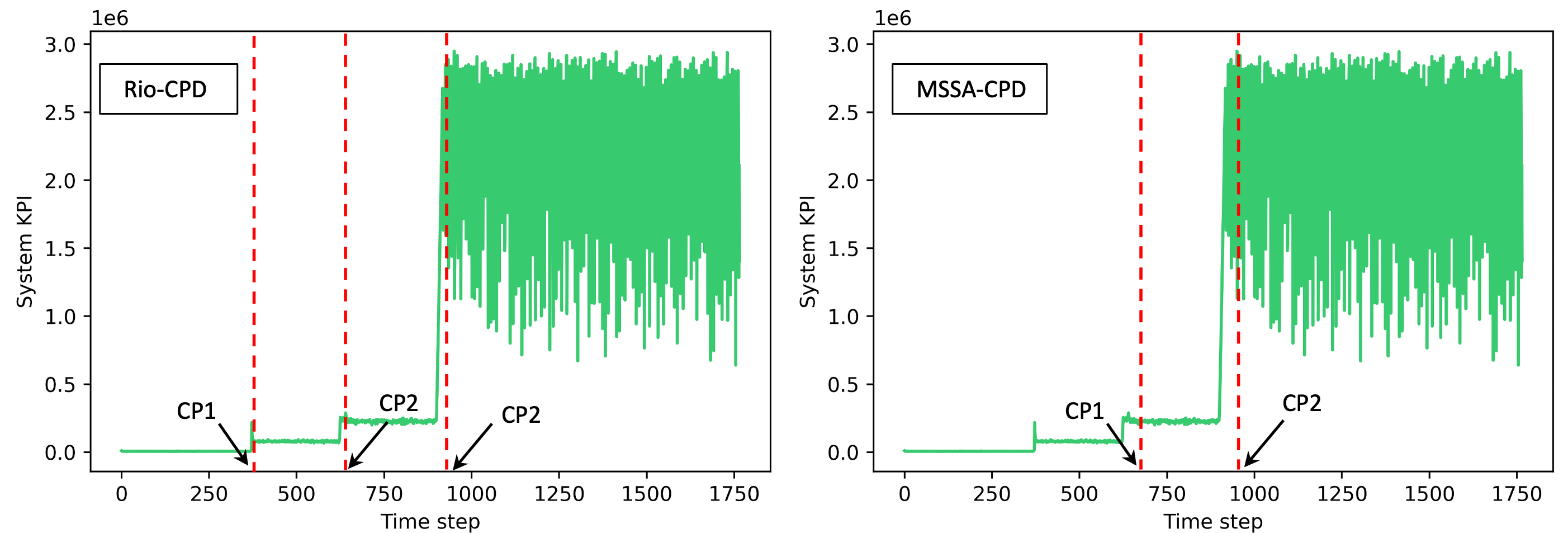}
    \caption{Detected change points by \cpd and MSSA-CPD on the Microservice dataset, false alarms omitted for MSSA-CPD.}
    \label{fig:cp}
\end{figure}

The results in~\Cref{tab:res-syn} indicate that \cpd, with both Log-Euclidean (LE) and Log-Cholesky (LC) metrics, consistently performs better than or at least competitively with other methods. In particular, for datasets with correlation-based change points, \cpd significantly outperforms other methods. We observe that other methods suffer from both false negatives and false positives—they tend to miss correlation changes while also reporting more change points due to local perturbations. Notably, KL-CPD has an F1 score of $0$ on the Microservice dataset due to the absence of any true positive detections. Furthermore, the LC metric is claimed to be more numerically stable and computationally efficient compared to the LE metric~\cite{lin2019riemannian}, which is supported by our experiments, as \cpd with the LC metric demonstrates slightly better performance. As an example, \Cref{fig:cp} illustrates the change points detected in the Microservice dataset with respect to the system key performance index (KPI), where \cpd promptly captures changes following KPI shifts.

\begin{table}[h!]
\centering
\tiny
\renewcommand{\arraystretch}{1.1}

\caption{Average delay on real-world and synthetic datasets. `N.A.' is used when no positive detection occurs within a reasonable delay window. All values are rounded to the nearest integer.}
\label{tab:ad}
\begin{adjustbox}{width=0.5\textwidth}
\begin{tabular}{c|ccc} 
\thickhline
Algorithm & {Microservice} & {Beedance} &  {HASC} \\
\hline
  KL-CPD & 11  & 7  & N.A. \\
  BOCPDMS & 8  & 6  & N.A. \\
  MSSA-CPD & 14  & 11  & 41  \\
  Contra-CPD  & 10  & 16  & 33 \\
  $\mathbf{\cpd}$ (\textbf{LE}) & 0  & 8  & 25  \\
  $\mathbf{\cpd}$ (\textbf{LC}) & 0  & 6  & 19  \\
 
\thickhline
  & {Connection} & {Speed} &  {Location} \\
\hline
  KL-CPD & N.A. & N.A.  & N.A. \\
  BOCPDMS & 9   & 6 &   7  \\
  MSSA-CPD & 2 & 4  & 3  \\
  Contra-CPD  & 5  & 5  & 4 \\
  $\mathbf{\cpd}$ (\textbf{LE})& 2  & 2  & 2 \\
  $\mathbf{\cpd}$ (\textbf{LC})& 2  & 2  & 3 \\
  \thickhline
\end{tabular}
\end{adjustbox}
\end{table}

\begin{figure}[h]
    \centering
    \includegraphics[width=\linewidth]{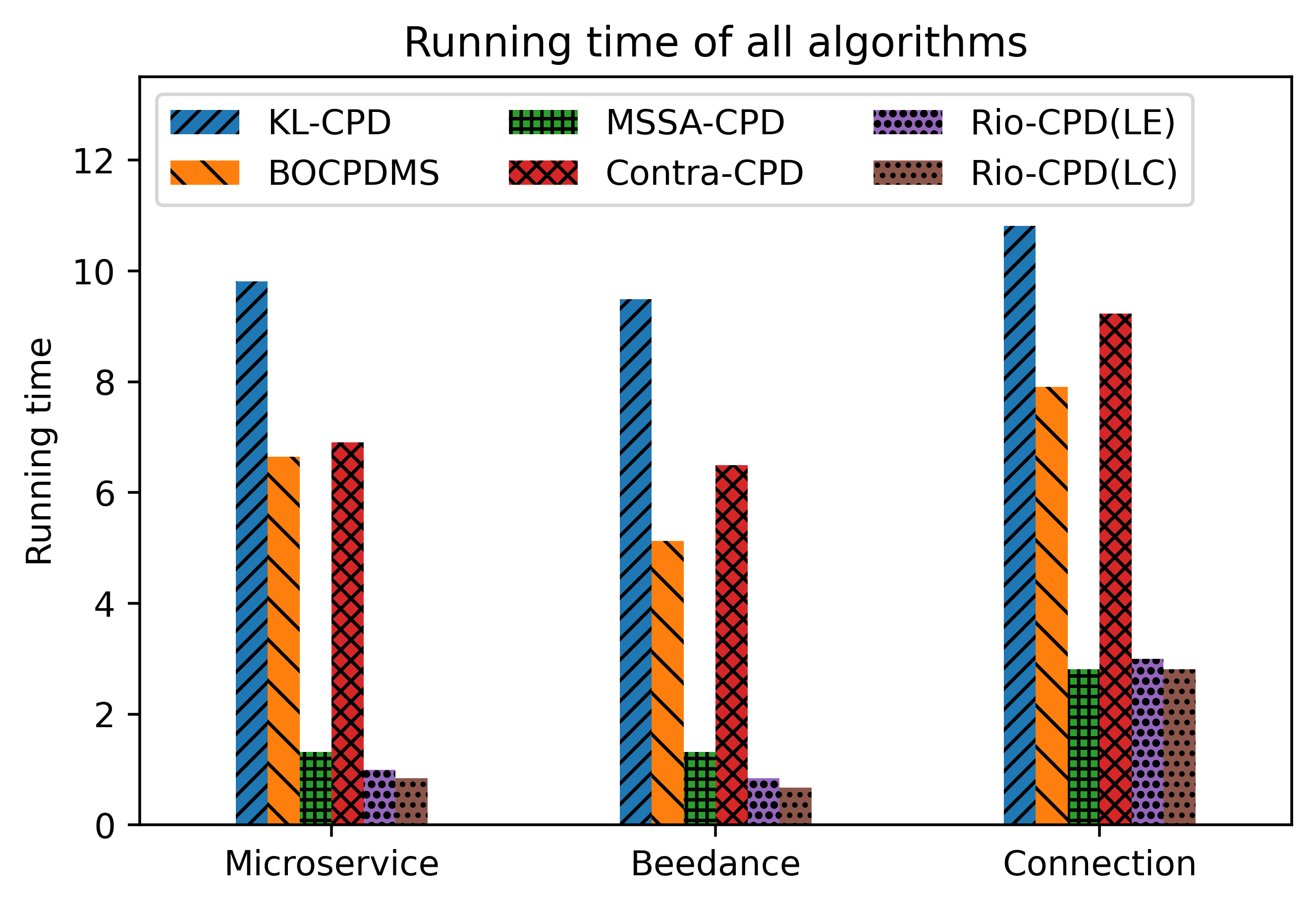}
    \vspace{-7mm}
    \caption{Running time comparison on three datasets. }
    \label{fig:time}
\end{figure}

We next evaluate the average detection delay for each algorithm. Detection delay measures the sensitivity of a CPD algorithm in an online setting, with a smaller delay being preferable. To avoid false alarms, we calculate the average delay only for points detected within twice the window size. Detections outside this range are not considered related to the current change point. The results are presented in~\Cref{tab:ad}. Additionally, we demonstrate the running time of all algorithms on three datasets: Microservice, Beedance, and Connection in \Cref{fig:time}. The reported numerical values are log-transformed, with the original running times provided in~\Cref{apdx:exp}.

An immediate observation is that \cpd, with either metric, runs faster than all baseline methods and has a lower detection delay. This efficiency is due to the simplicity of the metrics and our design of the CUSUM procedure with a closed-form solution for the Fr\'echet mean, which can be computed quickly.

\subsection{Human Action Recognition}



Human action recognition is another real-world application of change point detection, where a change point typically refers to abrupt changes in physical activity. We evaluate \cpd using both the Log-Euclidean and Log-Cholesky metrics on the WISDM dataset~\cite{weiss2019smartphone}, which contains 3-dimensional accelerometer measurements collected from a smartphone at a sampling rate of $20$ Hz. The dataset includes $17$ human activity changes, corresponding to $17$ change points. Following prior work~\cite{puchkin2023contrastive}, the dataset is pre-processed through sub-sampling.

We also compare \cpd against MSSA-CPD and Contra-CPD in terms of F1 score and detection delay, as previous results have consistently shown these methods outperform KL-CPD and BOCPDMS. Unless stated otherwise, the sliding window size for \cpd is set to $W=20$. The results are presented in~\Cref{tab:action}.

 \vspace{-3mm}
\begin{table}[h!]
\centering
\tiny
\renewcommand{\arraystretch}{1.1}

\caption{F1 score and average delay (AD) on the WISDM dataset.}
\vspace{0.1em}
\label{tab:action}
\begin{adjustbox}{width=0.5\textwidth}
\begin{tabular}{c|ccc} 
\thickhline
  Algorithm &  F1 (Default) & F1 (Best) & AD  \\
\hline
  MSSA-CPD & 0.225 &  0.406 & 15  \\
  Contra-CPD  & 0.314 & 0.378 & 23\\
  $\mathbf{\cpd}$ (\textbf{LE}) & 0.378 &  0.453 &  15 \\
  $\mathbf{\cpd}$ (\textbf{LC}) & \textbf{0.400} &  \textbf{0.495} &  14 \\
 
\thickhline
\end{tabular}
\end{adjustbox}
\end{table}


We again observe that \cpd, with both Riemannian metrics, significantly outperforms other methods, especially in terms of the F1 score. Moreover, the Log-Cholesky metric slightly outperforms the Log-Euclidean metric, as we have seen in \Cref{tab:res-syn}.





\section{Related Work}
\label{sec:related}

\noindent\textbf{Change Point Detection} has been studied in both offline and online settings. The \emph{offline CPD} problem is often approached as time series segmentation and has been extensively studied~\cite{chung2004evolutionary, lovric2014algoritmic, liu2008novel, truong2020selective, van2020evaluation}.
According to~\cite{aminikhanghahi2017survey}, change point detection methods can be categorized into four types: probabilistic, kernel-based, likelihood ratio, and subspace models. Likelihood ratio methods are typically parametric, while the others can be parametric or non-parametric.

\emph{Online CPD} has received more attention recently. The earliest notable method is BOCPD (Bayesian Online Change Point Detection)~\cite{adams2007bayesian}, which inspired numerous follow-up works~\cite{xuan2007modeling, malladi2013online, agudelo2020bayesian, alami2020restarted, knoblauch2018spatio}. Gaussian process-based methods~\cite{caldarelli2022adaptive} also fall under probabilistic models. Kernel-based approaches originated from kernel maximum mean discrepancy for change point detection~\cite{gretton2006kernel}, later refined for online settings~\cite{gong2012kernelized,li2019scan, li2015m}. Recently, neural network-based approaches~\cite{chang2019kernel,cheng2021neural,li2024automatic} have also been classified as kernel-based methods. On the other hand, likelihood ratio methods, with the longest history, began with CUSUM~\cite{page1954continuous} and the generalized likelihood ratio test~\cite{gustafsson1996marginalized}, followed by many variants~\cite{flynn2019change, wei2022online, romano2023fast, wang2024sequential, ward2023poisson}. These methods are parametric, assuming known distributions. Subspace models assume time series data lie in a low-dimensional manifold, approximated through singular spectrum analysis~\cite{alanqary2021change}, metric design~\cite{costa2006single, keriven2020newma, dubey2020frechet}, or other learning-based approaches~\cite{jiao2018subspace, kawahara2007change, fuh2020asymptotically}. Our \textsc{Rio-CPD} model falls into the category of refined subspace model based methods, but has merits on accuracy and efficiency by exploring the Riemannian geometry of correlation matrices. Also different from existing works, \cpd is capable of detecting both magnitude-based changes and correlation-based changes.


\paragraph{Riemannian Metrics on the Space of \  gspdn.} The manifold \spdn represents the space of positive definite matrices of dimension $n$, which can be equipped with various metrics. Canonical choices include the Log-Euclidean metric~\cite{arsigny2007geometric}, Log-Cholesky metric~\cite{lin2019riemannian}, affine-invariant metrics~\cite{moakher2005differential, pennec2006riemannian}, and Bures-Wasserstein metrics~\cite{dryden2009non,takatsu2011wasserstein,bhatia2019bures}. The Log-Euclidean and Log-Cholesky metrics exhibit Lie-group bi-invariance and provide simple closed forms for the Fr'echet average of SPD matrices. The affine-invariant metric is inverse-consistent and invariant under congruence actions, such as affine transformations, and is geodesically complete. The Bures-Wasserstein metric, while not geodesically complete, is bounded by the positive semi-definite cone. Recent works also explore the Riemannian geometry of correlation matrices~\cite{grubivsic2007efficient, david2019riemannian}, as well as more general positive semi-definite matrices~\cite{vandereycken2013riemannian, massart2020quotient}.

\section{Discussions}

%



\textbf{Limitations.} While \cpd leverages the Pearson correlation coefficient to capture linear relationships, it may not fully account for non-linear dependencies. Nonetheless, linear relationships often provide valuable practical insights. This trade-off is expected, as complex correlations typically require advanced techniques and complicate interpretability. Moreover, extending \cpd to high-dimensional data necessitates further exploration of suitable Riemannian structures, potentially enhancing the universality of the framework. We consider tackling these challenges in future work.

\textbf{Conclusion.} We address the novel problem of Correlation-aware Online Change Point Detection, which has become a crucial task in many real-world applications. We propose \cpd, a non-parametric framework inspired by the Riemannian geometry of correlation matrices. We instantiate \cpd with the Log-Euclidean and Log-Cholesky metrics, due to their simplicity and stability, and introduce a novel design of the CUSUM statistics. We evaluate \cpd using these metrics across various datasets, both with and without correlation changes, demonstrating its superior performance in terms of accuracy and efficiency.

\bibliography{reference}
\bibliographystyle{icml2024}

\newpage
\appendix

\section{More Details on Experiments}
\label{apdx:exp}
\subsection{Datasets}
\label{apdx:data}

The synthetic dataset is simulated using a physical particle-spring system, consisting of five particles moving within a rectangular space. Some randomly selected pairs of particles are connected by invisible springs. The motion of these particles is recorded as observations in a multivariate time series, capturing both their location and velocity. The movement of particles is governed by the principles of physics, such as Newton's law, Hooke's law, and the Markov property. The Connection dataset is modified by re-sampling certain pairs of particles to create new connections, thus satisfying the requirement for correlation changes, as discussed in~\cite{zhang2020correlation}. The other two datasets (Location and Velocity) contain change points caused by perturbations in location or velocity, which distinguishes them from the Connection dataset. Originally, these synthetic datasets were designed for a supervised offline setting, but we have adapted them for an unsupervised online setting in our experiments.

The Microservice dataset~\cite{zheng2024lemma} is generated from a microservice platform comprising six system nodes and 234 system pods. System faults were simulated on four different dates, resulting in 49 hours of recorded system activity. Specifically, six types of system metrics were collected (\textit{i.e.}, CPU usage, memory usage, received bandwidth, transmitted bandwidth, rate of received packets, and rate of transmitted packets), with a time granularity of 1 second. This dataset features correlation-based changes, as the recorded system metrics are influenced by the underlying relationships between system nodes and pods, which can potentially lead to system failure. The time points at which changes in system status occurred are recorded as change points.

\subsection{Experiments}



We provide further details on the experimental settings and results in this section. All experiments were conducted on a Linux system equipped with AMD EPYC 7302 16-core processors and two Nvidia RTX5500 GPUs with 24GB of memory, although most baseline methods do not require the use of a GPU.

Next, we outline the parameters used for \cpd. In the default settings shown in \Cref{tab:res-syn}, the parameters are adapted to different datasets but not specifically optimized. The parameters for baseline methods are inherited from prior implementations. Recall that \cpd has one "hard" parameter—the sliding window size—and two "soft" parameters: the threshold $\rho$ in CUSUM and $L$, which represents the lag between sampled data. For all datasets except HASC, we set $L=1$, as HASC contains a long sequence, for which we use $L=5$. The window size $W$ is set to 10 for the Beedance dataset, 20 for Microservice and HASC, and 5 for all synthetic datasets. For the CUSUM threshold, we tested a heuristic approach by setting $\rho$ to three times the variance of the pre-change distances, along with small integer values not greater than 5. The best results are reported based on the aforementioned strategy. As mentioned, the major parameter is the size of sliding window, which could vary depending on the length of the time series. Though our choice of $W$ most goes to 10 or 20, this may not be suitable for very short or lengthy time series.

In \Cref{tab:time-apdx}, we present the actual runtime of all algorithms on all datasets.

\begin{table}[h!]
\centering
\tiny
\renewcommand{\arraystretch}{1.2}

\caption{Computation time on real-world and synthetic datasets.}
\vspace{0.5em}
\label{tab:time-apdx}
\begin{adjustbox}{width=0.5\textwidth}
\begin{tabular}{c|ccc} 
\thickhline
  {Algorithm} & {Microservice} & {Beedance} &  {HASC} \\
\hline
  KL-CPD  & 15min  & 12min  & $>$1h \\
  BOCPDMS  &  100s  &  35s  & 10min \\
  MSSA-CPD  &  0.5s  & 0.5s   & 12s \\
  Contra-CPD  & 2min  & 90s  & 30min\\
  $\mathbf{\cpd}$ (\textbf{LE})  &  0.7s  & 0.5s  & 16s \\
  $\mathbf{\cpd}$ (\textbf{LC})  &  0.5s  & 0.5s  & 13s \\
 
\thickhline
  & {Connection} & {Speed} &  {Location} \\
\hline
  KL-CPD  & 30min  & 30min  & 30min \\
  BOCPDMS  & 4min   &  4min  &  4min \\
  MSSA-CPD  &  7s  &  7s  & 7s \\
  Contra-CPD  & 10min  & 9min  & 9min\\
  $\mathbf{\cpd}$ (\textbf{LE}) &  8s  &  8s & 8s\\
  $\mathbf{\cpd}$ (\textbf{LC}) &  7s  &  7s  & 7s\\
  \thickhline
\end{tabular}
\end{adjustbox}
\end{table}

\subsection{Discussions}
Geometry-inspired methods have recently gained attention in data mining and machine learning~\cite{le2004geometric, atz2021geometric}; however, their application to the CPD problem remains underexplored. In contrast, the Riemannian metric—a key concept in differential geometry—offers significant potential due to the broad applicability of symmetric positive definite (SPD) matrices, denoted as \spdn. Our framework is designed to accommodate all well-defined Riemannian metrics on \spdn. We have chosen the Log-Euclidean and Log-Cholesky metrics for integration into \cpd because of their efficiency and simplicity. Other metrics could be incorporated using the Karcher mean, a generalized version of the Fr\'echet mean, computed via Riemannian optimization methods. However, this approach may involve more complex techniques and could impact the efficiency or accuracy of the \cpd framework.

\end{document}